\documentclass[12pt]{article}
\usepackage{amsmath}
\usepackage{graphicx,psfrag,epsf}
\usepackage{enumerate}
\usepackage{natbib}
\usepackage{multirow}
\usepackage{afterpage}
\usepackage{rotating}
\usepackage{url} 
\usepackage{longtable}
\usepackage{caption}

\def\z{{\bf z}}

\begin{document}


  \title{\bf Deep Gaussian Mixture Models}
\author{Cinzia Viroli         \and
        Geoffrey J. McLachlan 
}
    \maketitle

\begin{abstract}
Deep learning is a hierarchical inference method formed by subsequent multiple layers of learning able to more efficiently describe complex relationships. In this work, Deep Gaussian Mixture Models are introduced and discussed. A Deep Gaussian Mixture model (DGMM) is a network of multiple layers of latent variables, where, at each layer, the variables follow a mixture of Gaussian distributions. Thus, the deep mixture model consists of a set of nested mixtures of linear models, which globally provide a nonlinear model able to describe the data in a very flexible way. In order to avoid overparameterized solutions, dimension reduction by factor models can be applied at each layer of the architecture thus resulting in deep mixtures of factor analysers.
\end{abstract}

\section{Introduction}
In the recent years, there has been an increasing interest on Deep Learning for supervised classification \citep{lecun}. It is very difficult to give an exact definition of what it is due to its wide applicability in different contexts and formulations, but it can be thought of as a set of algorithms able to gradually
learn a huge number of parameters in an architecture composed by
multiple non-linear transformations, called multi-layer structure. Deep Neural Networks have achieved great success in supervised classification and an important example of it is given by the so-called Facebook's DeepFace software: a deep learning facial recognition system
that employs a nine-layer neural network with over 120 million connection
weights. It can identify human faces in digital images with an accuracy of 97.35\%, at the same level as the human visual capability \citep{hodson}.
Deep learning architectures are now widely used for speech recognition, object detection, pattern recognition, image processing and many other supervised classification tasks; for a comprehensive historical survey and its applications, see \cite{schmidhuber} and the references therein.

Despite the success of deep models for supervised tasks, there has been limited research in the machine learning and statistics community on deep methods for clustering. In this paper we will present and discuss deep Gaussian mixtures for clustering purposes, a powerful generalization of classical Gaussian mixtures to multiple layers.
Identifiability of the model is discussed and an innovative stochastic estimation algorithm is proposed for parameter estimation.
Despite the fact that in recent years research on mixture models has been intense and prolific in many directions, we will show how deep mixtures can be very useful for clustering in complex problems.

The paper is organized as follows. In the next section classical Gaussian mixture models will be reviewed. In Section 3 deep Gaussian mixtures are defined and their main probabilistic properties presented. Identifiability is also discussed. In Section 4 dimensionally reduced deep mixtures are presented. Section 5 is devoted to the estimation algorithm for fitting the model. Experimental results on simulated and real data are presented in Section 6. We conclude this paper with some final remarks (Section 7).

\section{Gaussian Mixture Models}
Finite mixture models \citep{Peel} have gained growing popularity in the last decades as a tool for model-based clustering \citep{Fraley}. They are now widely used in several areas such as pattern recognition, data mining, image analysis, machine learning and in many problems involving clustering and classification methods.

Let $\textbf{y}_i$ be a $p$-dimensional random vector containing $p$ quantitative variables of interest for the statistical unit $i$th, with $i=1,\ldots,n$.
Then $\textbf{y}_i$ is distributed as a Gaussian Mixture Model (GMM) with $k$ components if

\begin{equation*}
f(\textbf{y}_i;\boldsymbol\theta) =
\sum_{j=1}^k\pi_j\phi^{(p)}(\textbf{y}_i;\mu_j,\Sigma_j),
\end{equation*}

\noindent where the $\pi_j$ are positive weights subject to $\sum_{j=1}^k\pi_j=1$ and the $\mu_j,\Sigma_j$ are the parameters of the Gaussian components. Note an interesting property that will be very useful in defining our proposal: a Gaussian mixture model has a related factor-analytic representation via a linear model with a certain prior probability as
\begin{equation*}
\textbf{y}_i=\mu_j + \Lambda_j \textbf{z}_i + \textbf{u}_i \ \ \ \ \textrm{with prob. } \pi_j,
\end{equation*}
\noindent where $\z_i$ is $p$-dimensional a latent variable with a multivariate standard Gaussian distribution and $\textbf{u}_i$ is an independent vector of random errors with $\textbf{u}_i\sim N(0,\Psi_j)$, where the $\Psi_j$ are diagonal matrices.  The component-covariance matrices can then be decomposed as $\Sigma_j=\Lambda_j \Lambda_j^\top+\Psi_j$.

\section{Deep Mixture Models}

Deep learning is a hierarchical inference method organized in a multilayered architecture, where the subsequent multiple layers of learning are able to efficiently describe complex relationships. In the similar perspective of deep neural networks, we define a Deep Gaussian Mixture model (DGMM) as a network of multiple layers of latent variables. At each layer, the variables follow a mixture of Gaussian distributions. Thus, the deep mixture model consists of a set of nested mixtures of linear models that globally provide a nonlinear model able to describe the data in a very flexible way.

\subsection{Definition}
Suppose there are $h$ layers. Given the set of observed data $\textbf{y}$ with dimension $n \times p$ at each layer a linear model to describe the data with a certain prior probability is formulated as follows:

\begin{eqnarray}\label{eqn1}
  &&(1) \ \ \ \ \ \ \ \ \textbf{y}_i=\eta_{s_1}^{(1)}+ \Lambda_{s_1}^{(1)} \textbf{z}_i^{(1)}+\textbf{u}_i^{(1)} \textrm{ with prob. } \pi_{s_1}^{(1)}, \ {s_1}=1,\ldots,k_1, \nonumber \\
  && (2)  \ \ \ \ \ \ \ \ \textbf{z}_i^{(1)}=\eta_{s_2}^{(2)}+ \Lambda_{s_2}^{(2)} \textbf{z}_i^{(2)}+\textbf{u}_i^{(2)} \textrm{ with prob. }  \pi_{s_2}^{(2)}, \ s_2=1,\ldots,k_2, \nonumber \\
 && \ \ \ \ \ \ \ \ \ \ \ \ \ ... \\
 && (h)  \ \ \ \ \ \ \ \ \textbf{z}_i^{(h-1)}=\eta_{s_h}^{(h)}+ \Lambda_{s_h}^{(h)} \textbf{z}_i^{(h)}+\textbf{u}_i^{(h)}   \textrm{ with prob. }   \pi_{s_h}^{(h)}, \ t=1,\ldots,k_h, \ \ \ \ \ \nonumber
\end{eqnarray}

\noindent where $\textbf{z}_i^{(h)}\sim N(\textbf{0},\textbf{I}_p)$ ($i=1,\ldots,n$) and $\textbf{u}_i^{(1)},\ldots,\textbf{u}_i^{(h)}$ are specific random errors that follow a Gaussian distribution with zero expectation and covariance matrices $\Psi_{s_1}^{(1)}, \ldots,\Psi_{s_h}^{(h)}$, respectively, $\eta_{s_1}^{(1)}, \ldots,\eta_{s_h}^{(h)}$ are vectors of length $p$, $\Lambda_{s_1}^{(1)}, \ldots,\Lambda_{s_h}^{(h)}$ are square matrices of dimension $p$. The specific random variables $\textbf{u}$ are assumed to be independent of the latent variables $\textbf{z}$. From this representation it follows that at each layer the conditional distribution of the response variables given the regression latent variables is a (multivariate) mixture of Gaussian distributions.

\begin{figure}
  \centering
  \includegraphics[scale=0.4]{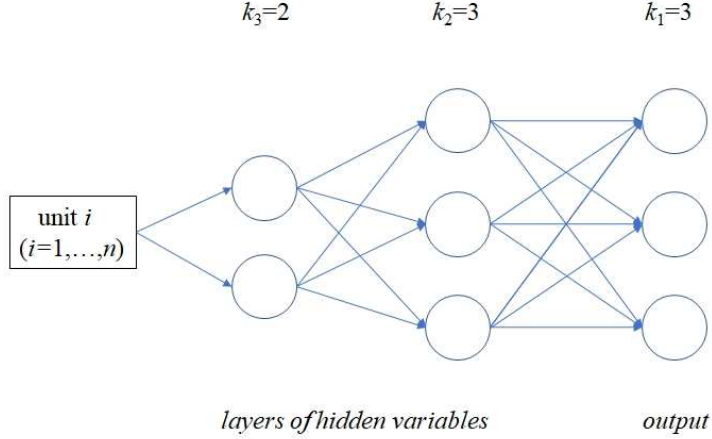}
  \caption{Structure of a DGMM with $h=3$ and number of layer components $k_1=3$, $k_2=3$ and $k_3=2$}\label{fig:example}
\end{figure}

To illustrate the DGMM, consider $h=3$ and let the number of layer components be $k_1=3$, $k_2=3$ and $k_3=2$. The structure is shown in Figure \ref{fig:example}. Thus, at the first layer we have that the conditional distribution of the observed data given $\z^{(1)}$ is a mixture with 3 components and so on. More precisely, by considering the data as the zero layer, $\textbf{y}=\textbf{z}^{(0)}$, all the conditional distributions follow a first order Markov first order property that is $f(\textbf{z}^{(l)}|\textbf{z}^{(l+1)},\textbf{z}^{(l+2)},\ldots,\textbf{z}^{(h)};\boldsymbol\Theta)=
f(\textbf{z}^{(l)}|\textbf{z}^{(l+1)};\boldsymbol\Theta)$ for $l=0,\ldots,h-1$. At each layer, we have
\begin{eqnarray}\label{cond}
f(\textbf{z}^{(l)}|\textbf{z}^{(l+1)};\boldsymbol\Theta)=\sum_{i=1}^{k_{l+1}}\pi_i^{(l+1)}N(\eta_i^{(l+1)}+
 \Lambda_i^{(l+1)} \textbf{z}^{(l+1)},\Psi_i^{(l+1)}).
\end{eqnarray}

\noindent Moreover, with the DGMM with $k_1=3$, $k_2=3$ and $k_3=2$ will have a `global' number of $M=8$ sub-components ($M=\sum_{l=1}^h \pi_l$), but final $k=18$ possible paths for the statistical units ($k=\prod_{l=1}^h \pi_l$) that share and combine the parameters of the $M$ sub-components. Thanks to this tying, the number of parameters to be estimated is proportional to the number of sub-components, thus reducing the computational cost to learning directly a model with $k=18$ components.

Let $\Omega$ be the set of all possible paths through the network.
The generic path $s=(s_1,\ldots,s_{h})$ has a probability $\pi_s$ of being sampled, with
$$ \sum_{s \in \Omega} \pi_s=\sum_{s_1,\ldots,s_{h}} \pi_{(s_1,\ldots,s_{h})}=1.$$

\noindent The DGMM can be written as

\begin{eqnarray}
  f(\textbf{y};\boldsymbol\Theta) &=& \sum_{s \in \Omega} \pi_s N(\textbf{y};\boldsymbol\mu_s,\boldsymbol\Sigma_s),
\end{eqnarray}

\noindent where

\begin{eqnarray*}
  \boldsymbol\mu_s&=& \eta_{s_1}^{(1)} + \Lambda_{s_1}^{(1)} (\eta_{s_2}^{(2)} +  \Lambda_{s_2}^{(2)}(\ldots (\eta_{s_{h-1}}^{(h-1)} +  \Lambda_{s_{h-1}}^{(h-1)}\eta_h^{(h)})))  \\
  &=&  \eta_{s_1}^{(1)} + \sum_{l=2}^h \left( \prod_{m=1}^{l-1} \Lambda_{s_m}^{(m)}\right) \eta_{s_l}^{(l)}
\end{eqnarray*}

\noindent and

\begin{eqnarray*}
  \boldsymbol\Sigma_s&=& \Psi_{s_1}^{(1)} + \Lambda_{s_1}^{(1)} (\Lambda_{s_2}^{(2)}(\ldots (\Lambda_{s_h}^{(h)}\Lambda_{s_h}^{(h)\top}+\Psi_{s_h}^{(h)})\ldots)\Lambda_{s_2}^{(2)\top})\Lambda_{s_1}^{(1)\top}  \\
  &=&  \Psi_{s_1}^{(1)} + \sum_{l=2}^h \left( \prod_{m=1}^{l-1} \Lambda_{s_m}^{(m)}\right) \Psi_{s_l}^{(l)}\left( \prod_{m=1}^{l-1} \Lambda_{s_m}^{(m)}\right)^\top.
\end{eqnarray*}

Thus globally the deep mixture can be viewed as a mixture model with $k$ components and a fewer number of parameters shared through the path. In a DGMM, not only the conditional distributions, but also the marginal distributions of the latent variables $\z^{(l)}$ are Gaussian mixtures. This can be established by integrating out the bottom latent variables, so that at each layer
\begin{eqnarray}\label{eqn:marg}
  f(\textbf{z}^{(l)};\boldsymbol\Theta) &=& \sum_{\tilde{s}= (s_{l+1},\ldots,s_h)} \pi_{\tilde{s}} N(\textbf{z}^{(l)};\tilde{\boldsymbol\mu}_{\tilde{s}}^{(l+1)},\tilde{\boldsymbol\Sigma}_{\tilde{s}}^{(l+1)}),
\end{eqnarray}

\noindent where $\tilde{\boldsymbol\mu}_{\tilde{s}}^{(l+1)}= \eta_{s_{l+1}}^{(l+1)} + \Lambda_{s_{l+1}}^{(l+1)} (\eta_{s_{l+2}}^{(l+2)} +  \Lambda_{s_{l+2}}^{({l+2})}(\ldots (\eta_{s_{h-1}}^{(h-1)} +  \Lambda_{s_{h-1}}^{(h-1)}\eta_h^{(h)})))$
and $ \tilde{\boldsymbol\Sigma}_{\tilde{s}}^{(l+1)}=\Psi_{s_{l+1}}^{(l+1)} + \Lambda_{s_{l+1}}^{(l+1)} (\Lambda_{s_{l+2}}^{(l+2)}(\ldots (\Lambda_{s_h}^{(h)}\Lambda_{s_h}^{(h)\top}+\Psi_{s_h}^{(h)})\ldots)\Lambda_{s_{l+2}}^{(l+2)\top})\Lambda_{s_{l+1}}^{(l+1)\top}$.

A deep mixture model for modeling natural images has been proposed by \cite{NIPS2014}. However, this model suffers from serious identifiability issues as discussed in the next section.

\subsection{Model-based clustering and identifiability}
As previously observed in a DGMM the total number of components (potentially identifying the groups) is given by the total number possible paths, $k$. In case the true number of groups, say $k^*$, is known, one could limit the estimation problem by considering only the models with $k_1=k^*$ ($k_1<k$) and perform clustering through the conditional distribution
$f(\textbf{y}|\textbf{z}^{(1)};\boldsymbol\Theta)$. This has the merit to have a nice interpretation: the remaining components of the bottom layers act as density approximations to the global non-Gaussian components. In this perspective, the model represents an automatic tool for merging mixture components \citep{Hennig2010,patrick,melnykov} and the deep mixtures can be viewed as a special mixture of mixtures model \citep{Li2005}.

However, in the general situation without further restrictions, the DGMM defined in the previous session suffers from serious identifiability issues related to the number of components at the different layers and the possible equivalent paths they could form. For instance, if $h=2$, a DGMM with $k_1=2$, $k_2=3$ components may be indistinguishable from a DGMM with $k_1=3$, $k_2=2$ components, both giving a total number of possible $k=6 \ (=k_1 \cdot k_2)$ paths.
Notice that even if $k^*$ is known and we fix $k_1=k^*$ there is still non-identifiability for models with more than two layers.

%

Moreover, in all cases, there is a serious second identifiability issue related to parameter estimation.

In order to address the first issue, the we introduce an important assumption on the model dimensionality: the latent variables at the different layers have progressively decreasing dimension, $r_1,r_2,\ldots,r_h$, where $p > r_1 > r_2 > \ldots, > r_h \geq 1$. As a consequence, the parameters at the different levels will inherit different dimensionality as well. This constraint has also the merit to avoid over-parameterized models, especially when $p$ is high.

The second identifiability issue arises from the presence of latent variables and it is similar in its nature to the identifiability issue that affects factor models. In particular, given an invertible matrix $A$ of dimension $r \times r$, with $r<p$, the factor model $y = \eta + \Lambda z +u$, with $u \sim N(0,\Psi)$, and the
transformed factor model $y = \eta + \Lambda A A^{-1}z+ u$ are indistinguishable, where $A$ is an orthogonal matrix and the factors have zero mean and identity covariance matrix. Thus there are
$r(r-1)/2$ fewer free parameters. This ambiguity can be avoided by imposing the constraint that $\Lambda^\top\Psi^{-1}\Lambda$ is diagonal with elements in decreasing order (see, for instance, \cite{Mardia}).

Moving along the same lines, in the DGMM, at each layer from 1 to $h-1$, we assume that the conditional distribution of the latent variables $f(\textbf{z}^{(l)}|\textbf{z}^{(l+1)};\boldsymbol\Theta)$ has zero mean and identity covariance matrix and the same diagonality constraint on the parameters at each level.


\section{Deep dimensionally reduced Gaussian mixture models}
Starting from the model (\ref{eqn1}), dimension reduction is obtained by considering layers that are sequentially described by latent variables with a progressively decreasing dimension, $r_1,r_2,\ldots,r_h$, where $p > r_1 > r_2 > \ldots, > r_h \geq 1$. The dimension of the parameters in (\ref{eqn1}) changes accordingly.

Consider as an illustrative example a two-layer deep model ($h=2$). In this case, the dimensionally reduced DGMM consists of the system of equations:
\begin{eqnarray*}\label{eqn2}
  &&(1) \ \ \  \textbf{y}_i=\eta_{s_1}^{(1)}+ \Lambda_{s_1}^{(1)} \textbf{z}_i^{(1)}+\textbf{u}_i^{(1)} \textrm{ with prob. }  \pi_{s_1}^{(1)}, \ j=1,\ldots,k_1,  \\
  && (2)  \ \ \ \textbf{z}_i^{(1)}=\eta_{s_2}^{(2)}+ \Lambda_{s_2}^{(2)} \textbf{z}_i^{(2)}+\textbf{u}_i^{(2)} \textrm{ with prob. } \pi_{s_2}^{(2)}, \ i=1,\ldots,k_2,
\end{eqnarray*}
where $\textbf{z}_i^{(2)}\sim N(\textbf{0},\textbf{I}_{r_2})$, $\Lambda_{s_1}^{(1)}$ is a (factor loading) matrix of dimension $p \times r_1$, $\Lambda_{s_2}^{(2)}$ has dimension $r_1 \times r_2$, and ${\Psi}_{s_1}^{(1)}$ and ${\Psi}_{s_2}^{(2)}$ are squared matrices of dimension $p \times p$ and $r_1 \times r_1$, respectively. The two latent variables have dimension $r_1$ and $r_2$, respectively with $p> r_1 >r_2 \geq 1$.

The model generalizes and encompasses several model-based clustering methods. Gaussian mixtures are trivially obtained in absence of any layer and dimension reduction.
Mixtures of factor analyzers \citep{MCLACHLAN2003} may be considered as a one-layer deep model, where $\Psi_{s_1}^{(1)}$ are diagonal and $\textbf{z}_i^{(1)}\sim N(\textbf{0},\textbf{I}_{r_1})$. When $h=2$ with $k_1=1$, ${\Psi}^{(1)}$ is diagonal, and $\Lambda_{s_2}^{(2)}=\{0\}$, the deep dimensionally reduced mixture coincides with mixtures of factor analyzers with common factor loadings \citep{Baek} and heteroscedastic factor mixture analysis \citep{Montanari2010}. The so-called mixtures of factor mixture analyzers introduced by \cite{Viroli2010} is a two-layer deep mixture with $k_1>1$, ${\Psi}_{s_1}^{(1)}$ diagonal and $\Lambda_{s_2}^{(2)}=\{0\}$.
Under the constraints that $h=2$, $\Psi_{s_1}^{(1)}$ and $\Psi_{s_2}^{(2)}$ are diagonal, the model is a deep mixture of factor analyzers \citep{Tang12}. In this work, the authors propose to learn one layer at a time. After estimating the parameters at each layer, samples from the posterior distributions for that layer are used as data for learning the next step in a greedy layer-wise learning algorithm. Despite its computational efficiency this multi-stage estimation process suffers from the uncertainty in the sampling of the latent variable generated values. A bias introduced at a layer will affect all the remaining ones and the problem grows with $h$, with the number of components and under unbalanced possible paths. In the next section we will present a unified estimation algorithm for learning all the model parameters simultaneously.


\section{Fitting Deep Gaussian Mixture Models}

Because of the hierarchical formulation of a deep mixture model, the EM algorithm represents the natural method for parameter estimation. The algorithm alternates between two steps and it consists of maximizing (M-step) and calculating the conditional expectation (E-step) of the complete-data log-likelihood function given the observed data, evaluated at a given set of parameters, say $\boldsymbol\Omega'$:
\begin{eqnarray}\label{eq:condexp}
&&E_{\textbf{z}^{(1)},\ldots,\textbf{z}^{(h)},\textbf{s}|\textbf{y}; \boldsymbol\Theta'}\left[ \log L_c(\boldsymbol\Theta)\right].
\end{eqnarray}

This implies that we need to compute the posterior distributions of the latent variables given the data in the E-step of the algorithm. In contrast to the classical GMM, where this computation involves only the allocation latent variable $\textbf{s}$ for each mixture component, in a deep mixture model the derivation of bivariate (or multivariate) posteriors is required, thus making the estimation algorithm very slow and not applicable to large data.

To further clarify this, consider the expansion of the conditional expectation in (\ref{eq:condexp}) as sum of specific terms. For a model with $h=2$ layers, it takes the following form

\begin{eqnarray}\label{eq:condexp2}
&&E_{\textbf{z},s|\textbf{y}; \boldsymbol\Theta'}\left[ \log L_c(\boldsymbol\Theta)\right]=
\sum_{s \in \Omega}\int
f(\textbf{z}^{(1)},s|\textbf{y};\boldsymbol\Theta') \log
f(\textbf{y}|\textbf{z}^{(1)},s;\boldsymbol\Theta)d\textbf{z}^{(1)} \nonumber \\
&+& \sum_{s \in \Omega}\int \int f(\textbf{z}^{(1)},\textbf{z}^{(2)},s|\textbf{y};\boldsymbol\Theta') \log
f(\textbf{z}^{(1)}|\textbf{z}^{(2)}, s;\boldsymbol\Theta)d\textbf{z}^{(1)} d\textbf{z}^{(2)} \nonumber \\
&+& \int f(\textbf{z}^{(2)}|\textbf{y};\boldsymbol\Theta') \log
f(\textbf{z}^{(2)}) d\textbf{z}^{(2)} + \sum_{s \in \Omega} f(s|\textbf{y};\boldsymbol\Theta') \log f(s;\boldsymbol\Theta).
\end{eqnarray}

A proper way to overcome these computational difficulties is to adopt a stochastic version of the EM algorithm (SEM), \citep{celeux} or its Monte Carlo alternative (MCEM) \citep{wei}. The principle underlying the handling of the latent variables is to draw observations (SEM) or samples of observations (MCEM) from the conditional density of the latent variables given the observed data, in order to simplify the computation of the E-step.

The strategy adopted is to draw pseudorandom observations at each layer of the network through the conditional density $f(\textbf{z}^{(l)}|\textbf{z}^{(l-1)},s;\boldsymbol \Theta')$, starting from $l=1$ to $l=h$, by considering as fixed, the variables at the upper level of the model for the current fit of parameters, where at the first layer $\textbf{z}^{(0)}=\textbf{y}$.


The conditional density $f(\textbf{z}^{(l)}|\textbf{z}^{(l-1)},s;\boldsymbol \Theta')$ can be expressed as

\begin{eqnarray}\label{eqn:inversa}
f(\textbf{z}^{(l)}|\textbf{z}^{(l-1)},s;\boldsymbol \Theta')=\frac{f(\textbf{z}^{(l-1)}|\textbf{z}^{(l)},s;\boldsymbol \Theta')f(\textbf{z}^{(l)}|s)}{f(\textbf{z}^{(l-1)}|s;\boldsymbol \Theta')},
\end{eqnarray}
\noindent where the denominator does not depend on $\textbf{z}^{(l)}$ and acts as a normalization constant, and the two terms in the numerator, conditionally on $s$, are Gaussian distributed according to equations (\ref{eqn:marg}) and (\ref{cond}):
\begin{itemize}
  \item[] $f(\textbf{z}^{(l-1)}|\textbf{z}^{(l)},s;\boldsymbol \Theta')=N(\eta_{s_l}^{(l)}+\Lambda_{s_l}^{(l)}\textbf{z}^{(l)},\Psi_{s_l}^{(l)})$,
  \item[] $f(\textbf{z}^{(l)}|s;\boldsymbol \Theta')=N(\tilde{\boldsymbol\mu}_{s_l}^{(l+1)},\tilde{\boldsymbol\Sigma}_{s_l}^{(l+1)})$.
 \end{itemize}

By substituting them in (\ref{eqn:inversa}), after some simple algebra, it is possible to show that
\begin{eqnarray}\label{eqn:inversa2}
f(\textbf{z}^{(l)}|\textbf{z}^{(l-1)},s)=N\left(\boldsymbol\rho_{s_l}(\textbf{z}^{(l-1)}),\boldsymbol\xi_{s_l}\right),
\end{eqnarray}
where
$$\boldsymbol\rho_{s_l}(\textbf{z}^{(l-1)})=\boldsymbol\xi_{s_l}\left(\left(\Lambda_{s_l}^{(l)}\right)^\top\left(\Psi_{s_l}^{(l)}\right)^{-1}(\textbf{z}^{(l-1)}-\boldsymbol\eta_{s_l}^{(l)})+\left(\tilde{\boldsymbol\Sigma}_{s_l}^{(l+1)}\right)^{-1}\tilde{\boldsymbol\mu}_{s_l}^{(l+1)}\right)$$
and $$\boldsymbol\xi_{s_l}=\left(\left(\tilde{\boldsymbol\Sigma}_{s_l}^{(l+1)}\right)^{-1}+\left(\Lambda_{s_l}^{(l)}\right)^\top\left(\Psi_{s_l}^{(l)}\right)^{-1}\Lambda_{s_l}^{(l)}\right)^{-1}.$$

This is the core of the stochastic perturbation of the EM algorithm. Due to the sequential hierarchical structure of the random variable generation, the E and M steps of the algorithm can be computed for each layer. Considering the sample of $n$ observations, at the layer $l=1,\ldots,h$, we maximize
\begin{eqnarray}\label{eq:M1}
&& E_{\textbf{z}^{(l)},\textbf{s}|\textbf{z}^{(l-1)};\boldsymbol\theta'}\left[\sum_{i=1}^n \log f(\textbf{z}^{(l-1)}_i|\textbf{z}_i^{(l)},s;\boldsymbol\Theta)\right] \nonumber
\\ &=& \sum_{i=1}^n \int f(\textbf{z}_i^{(l)},s|\textbf{z}^{(l-1)}_i;\boldsymbol\Theta')\log f(\textbf{z}^{(l-1)}_i|\textbf{z}_i^{(l)},s;\boldsymbol\Theta)
d\textbf{z}_i
\end{eqnarray}
with respect to $\Lambda_{s_l}^{(l)}$, $\Psi_{s_l}^{(l)}$, and $\eta_{s_l}^{(l)}$. By considering
$f(\textbf{z}^{(l-1)}|\textbf{z}^{(l)},s)=N(\eta_{s_l}^{(l)}+\Lambda_{s_l}{(l)}\textbf{z}^{(l)},\Psi_{s_l}^{(l)}$), we can compute the score of (\ref{eq:M1}) to derive the estimates for the new parameters given the provisional ones. Therefore, the complete stochastic EM algorithm can be schematized as follows. For $l=1,\dots,h$:

\bigskip

\noindent\rule[0.5ex]{\linewidth}{0.7pt}
{\footnotesize
\begin{description}
  \item[-] S-STEP ($\textbf{z}_i^{(l-1)}$ is known) \\
  Generate $M$ replicates $\textbf{z}_{i,m}^{(l)}$ from $f(\textbf{z}_i^{(l)}|\textbf{z}_i^{(l-1)},s;\boldsymbol\Theta')$.
  \item[-] E-STEP - Approximate: \\
  $$E[\textbf{z}_i^{(l)}|\textbf{z}_i^{(l-1)},s;\boldsymbol\Theta'] \cong \frac{\sum_{m=1}^M \textbf{z}_{i,m}^{(l)}}{M}$$ and $$E[\textbf{z}_i^{(l)}\textbf{z}_i^{(l)\top}|\textbf{z}_i^{(l-1)},s;\boldsymbol\Theta'] \cong \frac{\sum_{m=1}^M \textbf{z}_{i,m}^{(l)}\textbf{z}_{i,m}^{(l)\top}}{M}.$$
  \item[-] M-STEP - Compute:\\
\begin{eqnarray*}
\hat{\Lambda}_{s_l}^{(l)}&=&\frac{\sum_{i=1}^np(s|\textbf{z}_i^{(l-1)})(\textbf{z}_i^{(l-1)}-\eta_{s_l}^{(l)})E[\textbf{z}_i^{(l)\top}|\textbf{z}_i^{(l-1)},s]
E[\textbf{z}_i^{(l)}\textbf{z}_i^{(l)\top}|\textbf{z}_i^{(l-1)},s]^{-1}}
{\sum_{i=1}^np(s|\textbf{z}_i^{(l-1)})},\\
\hat{\Psi}_{s_l}^{(l)}&=&\frac{\sum_{i=1}^np(s|\textbf{z}_i^{(l-1)})\left[(\textbf{z}_i^{(l-1)}-\eta_{s_l})(\textbf{z}_i^{(l-1)}-\eta_{s_l})^\top-(\textbf{z}^{(l-1)}_i-\eta_{s_l})
E[\textbf{z}_i^{(l)\top}|\textbf{z}_i^{(l-1)},s]\hat{\Lambda}_{s_l}^\top\right]} {\sum_{i=1}^np(s|\textbf{z}_i^{(l-1)})},\\
\hat{\eta}_{s_l}^{(l)}&=&\frac{\sum_{i=1}^n p(s_|\textbf{z}_i^{(l-1)})\left[\textbf{z}_i^{(l-1)}-\Lambda_{s_l}
E[\textbf{z}_i^{(l)\top}|\textbf{z}_i^{(l-1)},s]\right]} {\sum_{i=1}^np(s|\textbf{z}_i^{(l-1)})}, \\
\hat\pi_{s}^{(l)}&=&\sum_{i=1}^n f\left(s_l|\textbf{y}_i\right),
\end{eqnarray*}
\end{description}}

\noindent where $f\left(s_l|\textbf{y}_i\right)$ is the posterior probability of the allocation variable given the observed data that can be computed via Bayes' formula.


\section{Simulated and Real Application}
\subsection{Smiley Data}
\begin{figure}
  \centering
  \includegraphics[scale=0.7]{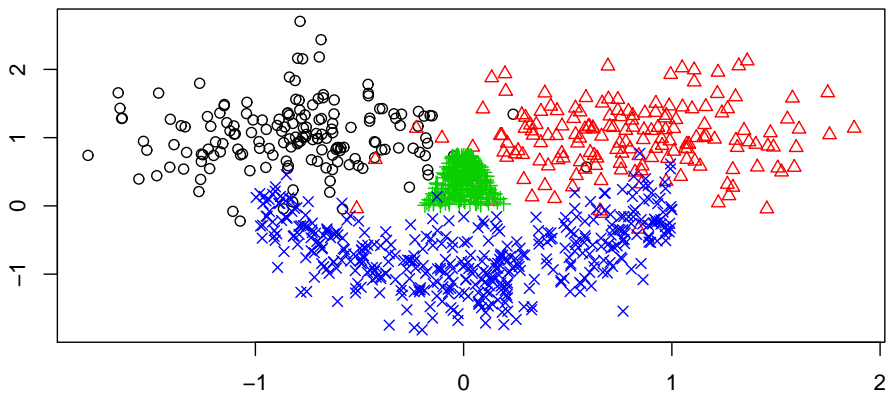}
  \caption{Smiley Data}\label{fig2}
\end{figure}

In this simulation experiment we have generated $n=1000$ observations from four classes in 3-dimensional space. The first two variables are relevant for clustering and have been generated by using the R package \verb"mlbench". They are structured into two Gaussian eyes, a triangular nose and a parabolic mouth, as shown in Figure \ref{fig2}. We have taken the standard deviation for eyes and mouth equal to 0.45 and 0.35, respectively. The third variable is a noise variable, independently generated from a Gaussian distribution with standard deviation 0.5.

Data have been independently generated 100 times. On each replicate, we applied DGMM with two-layers with $r_1=2$, $r_2=1$, $k_1=4$, and $k_2$ ranging from 1 to 5. We fitted the models 10 times in a multistart procedure and we selected the best fit according to BIC.

We compared the DGMM results with several clustering methods by fixing the number of groups equal to the true $k=4$ for all strategies. We fitted a Gaussian Mixture Model (GMM) by using the \verb"R" package \verb"Mclust" \citep{mclust1}, skew-normal and skew-t Mixture Models (SNmm and STmm) by using the \verb"R" package \verb"EMMIXskew" \citep{wang}, $k$-means, Partition around Medoids (PAM), and by Ward's method (Hclust) implemented hierarchically. Clustering performance is measured by the Adjusted Rand Index (ARI) and the misclassification rate. The average of the two indicators across the 100 replicates together with their standard errors are reported in Table \ref{tab1}.

\begin{table}[t]
\caption{Results on Smiley datasets: average of Adjusted Rand Index and misclassification rates across the 100 replicated. Standard errors are reported in brackets.\label{tab1}}
\begin{center}
\begin{tabular}{|l|cc|}
  \hline
  Method & ARI& m.r. \\
   \hline
  k-means & 0.661 \ \ (0.003) & \ \ 0.134 \ \ (0.001)\\
  PAM & 0.667 \ \ (0.004) & \ \ 0.132 \ \ (0.001)\\
  Hclust & 0.672 \ \ (0.013) & \ \ 0.141 \ \ (0.006)\\
    GMM  & 0.653 \ \ (0.008) & \ \ 0.178 \ \ (0.006)\\
    SNmm & 0.535 \ \ (0.006) & \ \ 0.251 \ \ (0.006)\\
    STmm & 0.566 \ \ (0.006) & \ \ 0.236 \ \ (0.004)\\
    DGMM & 0.788 \ \ (0.005) & \ \ 0.087 \ \ (0.002)\\
  \hline
\end{tabular}
\end{center}
\end{table}

\begin{figure}
  \centering
  \includegraphics[scale=0.6]{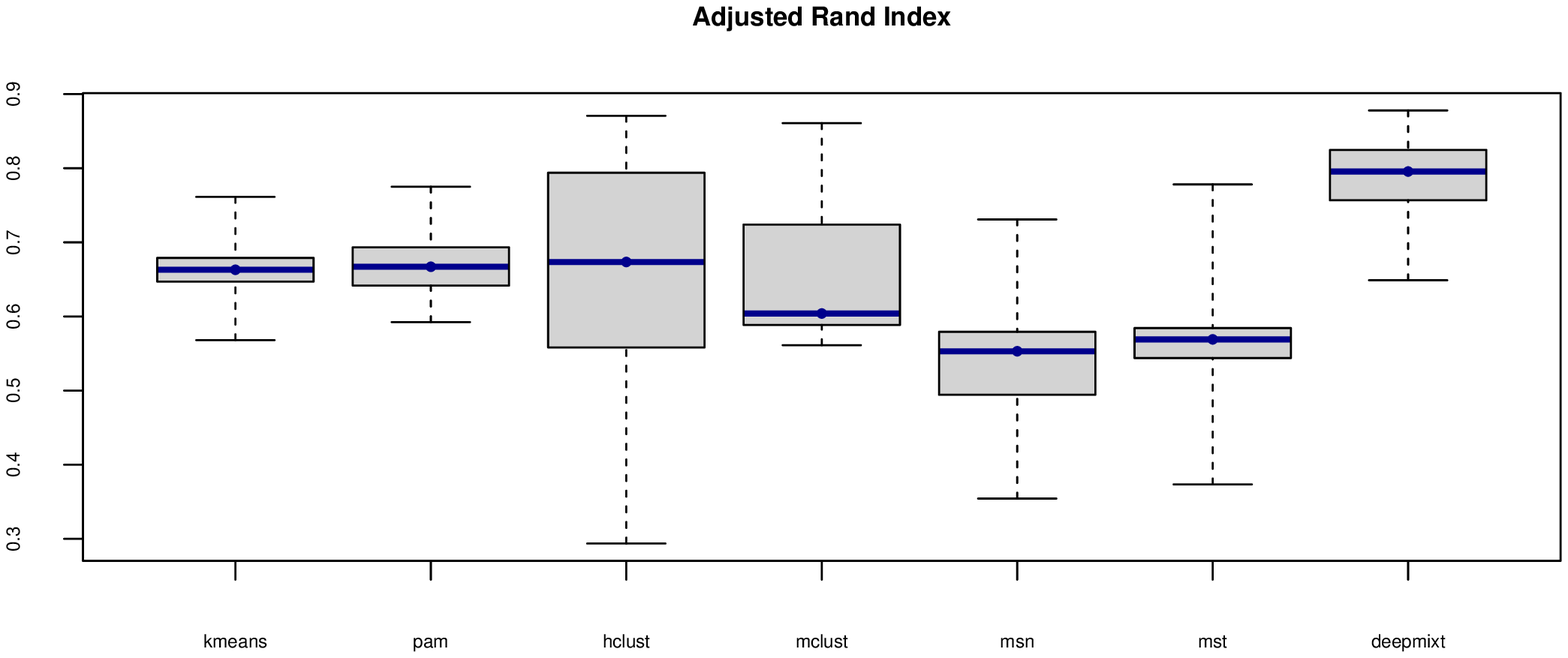}\\
  \includegraphics[scale=0.6]{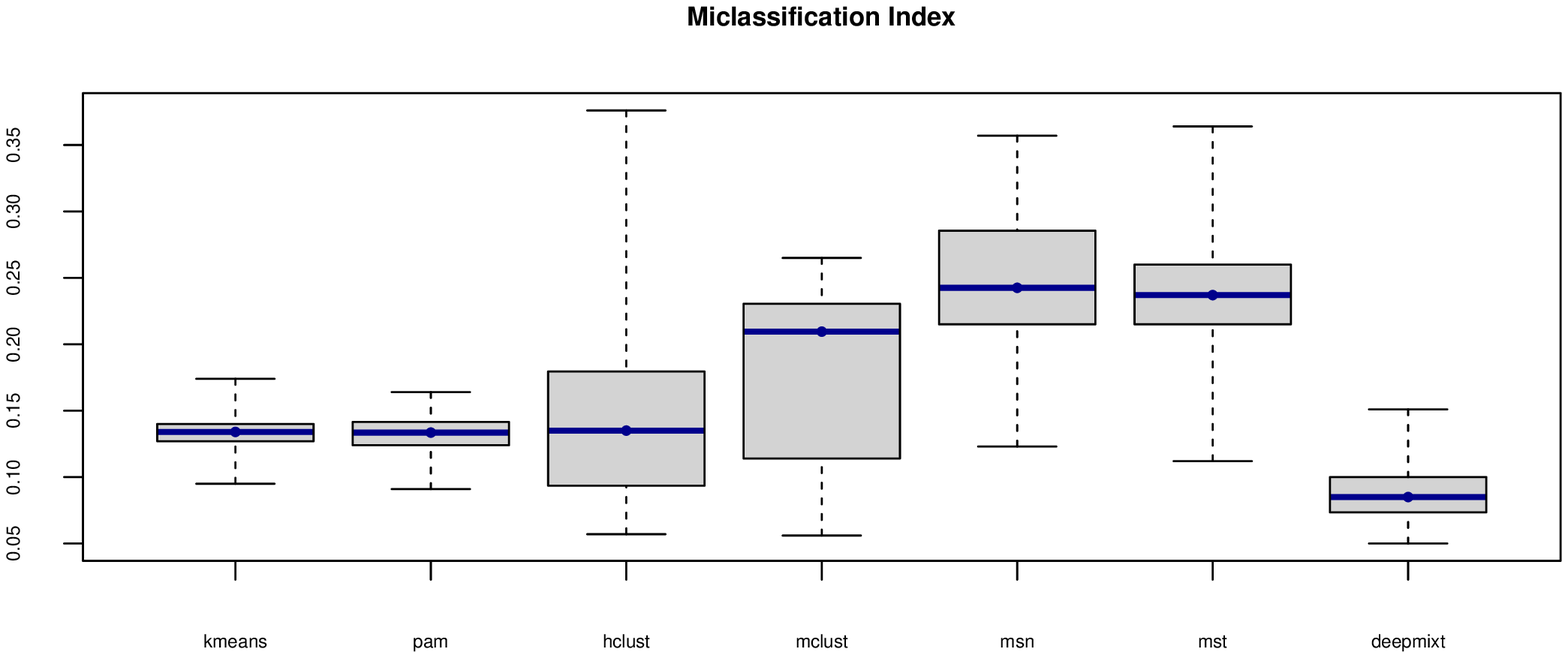}
  \caption{Smiley Data: Box plots of the Adjusted Rand Indices and Miclassification rates across the 100 replicates.}\label{fig3}
\end{figure}

Figure \ref{fig3} shows the box plots of the Adjusted Rand Indices and miclassification rates (m.r.'s) across the 100 replicates.
The results indicate that DGMM achieves the best classification performance compared to the other methods.

\subsection{Real Data}
In this section we shall apply the deep mixture model to some benchmark data used by the clustering and classification community.
We shall consider:
\begin{itemize}
  \item \emph{Wine Data}: this dataset comes from a study \citep{forina} on 27 chemical and physical properties of three types of wine
from the Piedmont region of Italy: Barolo (59), Grignolino (71), and Barbera (48). The clusters are well separated and most clustering methods give high clustering performance on this data. 
   \item \emph{Olive Data}: The dataset contains the percentage composition of eight fatty acids found by lipid fraction of 572 Italian olive oils \citep{forina2}. The data come from three regions: Southern Italy (323), Sardinia (98), and Northern Italy (151) and the aim is to distinguish between them. Also in this case, the clustering is not a very difficult task even if the clusters are not balanced. 
  \item \emph{Ecoli Data}: data consist of $n=336$ proteins classified into their various cellular localization sites based on
their amino acid sequences. There are $p=7$ variables and $k=8$ really unbalanced groups that make the clustering task rather difficult: cp cytoplasm (143), inner membrane
without signal sequence (77), perisplasm (52), inner membrane,
uncleavable signal sequence (35), outer membrane (20), outer
membrane lipoprotein (5), inner membrane lipoprotein (2), inner
membrane, cleavable signal sequence (2). These data are available from the UCI machine learning repository.
  \item \emph{Vehicle Data}: the dataset contains $k=4$ types of vehicles: a double decker bus (218), Cheverolet van (199), Saab 9000 (217) and an Opel Manta 400 (212). The aim is to cluster them on the basis of their silhouette represented from many different angles for a total of $p=18$ variables. This is a difficult classification task. In particular, the bus, the van and the cars are distinguishable, but it is very difficult to distinguish between the cars. The data are taken from the R library \verb"mlbench".
        \item \emph{Satellite Data}: the data derive from multi-spectral, scanner images purchased from NASA by the
Australian Centre for Remote Sensing. They consist of 4 digital images of the same scene in different spectral bands
  structured into $3 \times 3$ square neighborhood of pixels. Therefore, there are $p=36$ variables. The number of images is $n=6435$ coming from $k=6$ groups of images: red soil (1533), cotton crop (703), grey
soil (1358), damp grey soil (626), soil with vegetation stubble (707) and very damp grey soil (1508). This is notoriously a difficult clustering task not only because there are 6 unbalanced classes, but also because classical methods may suffer from the dimensionality $p=36$. The data are available from the UCI machine learning repository.

\end{itemize}

 On these data we compared the DGMM model with Gaussian Mixture Models (GMM), skew-normal and skew-t Mixture Models (SNmm and STmm), $k$-means and the Partition Around Medoids (PAM), hierarchical clustering with Ward distance (Hclust), Factor Mixture Analysis (FMA), and Mixture of Factor Analyzers (MFA). For all methods, we assumed the number of groups to be known. This assumption is made in order to compare the respective clustering performances. Note that in the case of an unknown number of groups, model selection for the DGMM can be done similarly to all the other mixture based approaches by using information criteria. Therefore, we considered the DGMM with $h=2$ and $h=3$ layers, a number of subcomponents in the hidden layers ranging from 1 to 5 (while $k_1=k^*$) and all possible models with different dimensionality for the latent variables under the constraint $p>r_1> ... >r_h\geq 1$. Moreover, we considered 10 different starting points for all possible models.
 For the GMM we considered all the possible submodels according to the family based on the covariance decomposition implemented in \verb"mclust". Finally, we fitted FMA and MFA by using the R package \verb"MFMA" available from the first author's webpage with different starting points and different number of latent variables ranging from 1 to the maximum admissible number.

In all cases we selected the best model according to BIC.

\begin{table}[t]
\small
\caption{Results on Real Data: Adjusted Rand Index (ARI) and misclassification rates (m.r.). \label{tab2}}
\begin{center}
\begin{tabular}{|l|cc|cc|cc|cc|cc|}
  \hline
  \emph{Dataset} & \multicolumn{2}{|c|}{\emph{Wine}} & \multicolumn{2}{|c|}{\emph{Olive}} & \multicolumn{2}{|c|}{\emph{Ecoli}} & \multicolumn{2}{|c|}{\emph{Vehicle}} & \multicolumn{2}{|c|}{\emph{Satellite}}\\
   & ARI & m.r. & ARI & m.r. & ARI & m.r. & ARI & m.r. & ARI & m.r. \\
   \hline
  $k$-means & 0.930 & 0.022 & 0.448 & 0.234 & 0.548 & 0.298 & 0.071 & 0.629 & 0.529 & 0.277 \\
     PAM  & 0.863 & 0.045 & 0.725 & 0.107 & 0.507 & 0.330 & 0.073 & 0.619 & 0.531 & 0.292 \\
  Hclust  & 0.865 & 0.045 & 0.493 & 0.215 & 0.518 & 0.330 & 0.092 & 0.623 & 0.446 & 0.337 \\
    GMM   & 0.917 & 0.028 & 0.535 & 0.195 & 0.395 & 0.414 & 0.089 & 0.621 & 0.461 & 0.374 \\
    SNmm  & 0.964 & 0.011 & 0.816 & 0.168 & -     & -     & 0.125 & 0.566 & 0.440 & 0.390 \\
    STmm  & 0.085 & 0.511 & 0.811 & 0.171 & -     & -     & 0.171 & 0.587 & 0.463 & 0.390 \\
    FMA   & 0.361 & 0.303 & 0.706 & 0.213 & 0.222 & 0.586 & 0.093 & 0.595 & 0.367 & 0.426 \\
    MFA   & 0.983 & 0.006 & 0.914 & 0.052 & 0.525 & 0.330 & 0.090 & 0.626 & 0.589 & 0.243 \\
    DGMM  & 0.983 & 0.006 & 0.997 & 0.002 & 0.749 & 0.187 &  0.191 & 0.481 & 0.604 & 0.249 \\
  \hline
\end{tabular}
\end{center}
\end{table}

For the smaller dataset (\emph{Wine}, \emph{Olive}, \emph{Ecoli}, \emph{Vehicle}) the best DGMM suggested by BIC was the model with $h=2$ layers, while $h=3$ layers were suggested for the \emph{Satellite} data.
The \emph{Wine} data are quite simple to classify. Most methods performed quite well. The best DGMM model was obtained with $r_1=3, \ r_2=2$ and $k_1=3, \ k_2=1$.
The \emph{Olive} data are not very well distinguished by classical methods such as $k$-means and hierarchical clustering, while model-based clustering strategies produce better performance.
Here deep learning with $r_1=5, \ r_2=1$ and $k_1=3 \ k_2=1$ suggested by BIC, gives excellent results with only 1 misclassified unit.

The challenging aspect of a cluster analysis on \emph{Ecoli} data is the high number of (unbalanced) classes. On these data SNmm and STmm did not reach convergence due to their being unable to handle satisfactorily the presence of two variables that each took on only two distinct values. The best clustering method also in this case is given by the deep mixture with $r_1=2, \ r_2=1$ and $k_1=8, \ k_2=1$.

Deep mixtures performed better than the other methods also for the difficult task to distinguish between silhouettes of \emph{vehicles} with progressively dimension reduction of $r_1=7, \ r_2=1$ and components $k_1=4, \ k_2=3$.

Finally, for the \emph{Satellite} data a DGMM with $h=3$ layers and $r_1=13, \ r_2=2, \ r_1=1$ and $k_1=6, \ k_2=2, \ k_1=1$ is preferred in terms of BIC. Results here are comparable with MFA with 4 factors, its having slightly higher ARI but with less corrected classified units in the total.

\section{Final remarks}
In this work a deep Gaussian mixture model (DGMM) for unsupervised classification has been investigated. The model is a very general framework that encompasses classical mixtures, mixtures of mixtures models, and mixture of factor analyzers as particular cases. Since DGMM is a generalization of classical model-based clustering strategies, it is guaranteed to work as well as these methods. We demonstrate the greater flexibility of DGMM with its higher complexity; for this reason it is particularly suitable for data with large sample size. 

We illustrated the model on simulated and real data. From the experimental study we conducted, the method works efficiently and it gives a good clustering performance with $h=2$ and $h=3$ layers where, as suggested, model choice can be undertaken according to information criteria.



\bibliographystyle{Chicago}

\bibliography{ref}

\end{document}